\documentclass{opt2026} 

\usepackage{float}
\makeatletter
\AtBeginDocument{\renewcommand*{\@jmlrenddoc}{%
  \phantomsection
  \protected@edef\@currentlabelname{end of \@shorttitle}%
  \label{jmlrend}%
  \global\let\@reprint\@empty}}
\makeatother
\usepackage{booktabs}
\usepackage{array}
\usepackage{tikz}
\usetikzlibrary{positioning}
\newcommand{\harnessA}{Harness~A}
\newcommand{\harnessB}{Harness~B}
\newcommand{\candidateC}{Candidate~C}

\title[Building the Harness Automatically]{Code-to-Harness: Distilling Black-Box Optimizers from Self-Play}

\optauthor{%
  \Name{Yi Wu$^*$} \Email{wuyish@google.com}\\
  \Name{Zheng Ren$^*$} \Email{murphyren@google.com}\\
  \Name{Zhiyu Hu} \Email{harryhu@google.com}\\
  \Name{Haochen Wang} \Email{haochenww@google.com}\\
  \Name{Daryl Chang} \Email{dlchang@google.com}\\
  \Name{Li Wei} \Email{liwei@google.com}\\
  \Name{Ting Wang} \Email{tingwa@google.com}\\
  \Name{Zhen Li} \Email{zhenlizh@google.com}\\
  \Name{Pooja Gupta} \Email{poojagupta@google.com}\\
  \Name{Nitin Jindal} \Email{nitinjindal@google.com}\\
  \Name{Lukasz Heldt} \Email{heldt@google.com}\\
  \addr Google \\
  Mountain View, USA
}

\begin{document}

\maketitle
\def\thefootnote{*}\footnotetext{Equal contribution to the work.}\def\thefootnote{\arabic{footnote}}

\begin{abstract}
Can an agent learn a numerical search strategy through executable practice and then transfer that strategy as text?  We study low-budget black-box optimization, where unaided language models remain well below strong classical optimizers.  During development, an agent repeatedly writes and evaluates optimizer programs.  It then distills the resulting program and practice record once into a 197-word primary harness (\harnessA{}), which is frozen before evaluation.  \harnessA{}  reduces Gemini Flash regret by 48\% in an independent $N=30$ study ($p<.001$), enters the GP-BO performance range on the practice family, and lowers mean regret on all three held-out BBOB landscapes.  The same text improves every tested Gemini executor and transfers to Claude Sonnet, reducing regret by 43\% and 49\% ($p\leq.005$).  An independent end-to-end replication produces \harnessB{}, a different program and text at the same performance tier. The same framework also attains the lowest regret on a sealed YouTube reward-tuning production benchmark. Executable practice is thus a viable way to discover a search policy, and language a portable medium for deploying it.\footnote{Code: \url{https://anonymous.4open.science/r/llm-opt-sandbox-E5B3/README.md}.}
\end{abstract}

\section{Introduction}
\label{sec:introduction}

Many scientific and engineering tasks require optimizing an expensive,
gradient-free objective under a small query budget --- the setting of Bayesian
optimization (BO) and evolutionary strategies
\citep{frazier2018tutorial,hansen2021coco}.  LLMs can interpret optimization
histories and express search heuristics, but are unreliable numerical optimizers
without structure \citep{huang2024truepotential}; recent work therefore embeds
them inside established search procedures \citep{liu2024llambo,agarwal2025bopro}.

Our question is how to learn search behavior that remains dependable after
development.  Prompt optimizers revise instructions from feedback
\citep{yang2024opro,pryzant2023protegi,wang2024promptagent}, and reflective
systems extend this idea to harness evolution
\citep{agrawal2026gepa,lee2026metaharness}.  Revising from test outcomes,
however, leaks evaluation information.  We instead distill practice on external
development objectives once, gate it, and freeze it before public or internal
evaluation; accepted runs are tested fresh and rejected runs are audited
separately.

\paragraph{Contributions.}
We contribute (i) a practice--gate--distill--freeze framework for reusable
low-budget search behavior that transfers unchanged across executors; and (ii)
an evaluation that seals before testing, reproduces one independently accepted
run, and exposes a rejected run.

\begin{figure}[!b]
  \centering
  \includegraphics[width=\linewidth,trim=0 10 0 10,clip]{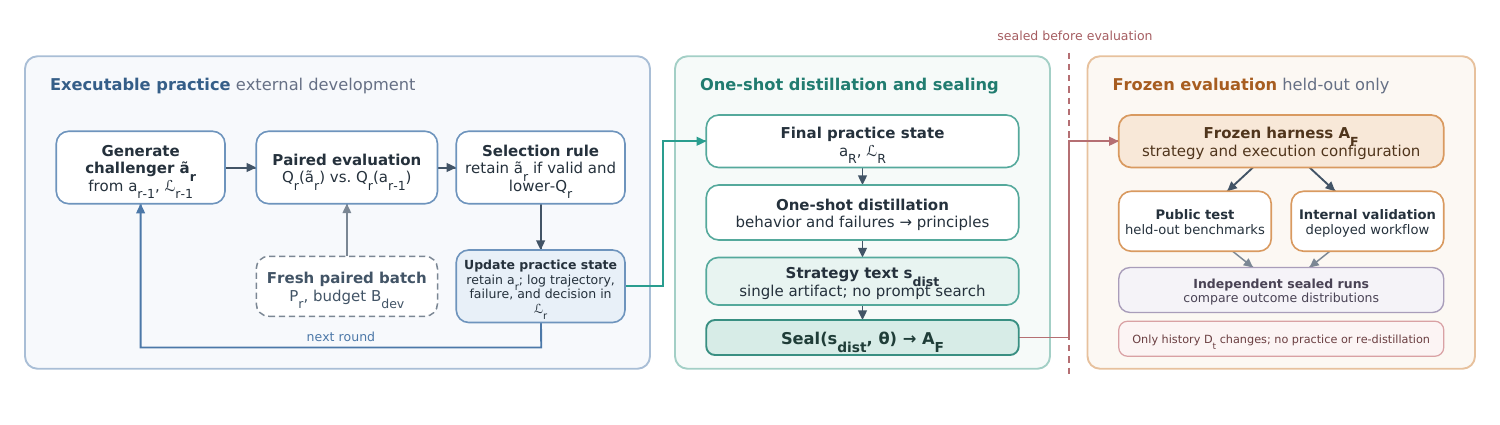}
  \caption{Method overview.  Repeated practice produces a champion and record;
  only a run that passes the sealed development gate is distilled and sealed as
  $\mathcal{A}_{\mathrm F}$, which remains fixed during public and internal
  evaluation.}
  \label{fig:method-overview}
\end{figure}


\section{Related Work}
\label{sec:related-work}

\paragraph{LLMs for black-box optimization.}
OPRO uses textual trajectories to propose solutions \citep{yang2024opro};
LLAMBO and BOPRO couple language models with BO components and
exploration--exploitation control \citep{liu2024llambo,agarwal2025bopro}.  We
instead learn a reusable policy across external objectives.

\paragraph{LLMs that write optimizers.}
FunSearch, AlphaEvolve, and LLaMEA evolve optimizer code
\citep{romera2024funsearch,novikov2025alphaevolve,vanstein2024llamea}; LLaMEA-BO
extends this to BBOB \citep{li2025llameabo}.  Our practice stage follows this
paradigm in miniature, but distills experience back into portable language.

\paragraph{Feedback-based policy improvement.}
Feedback can drive critique, tree search, reflection, genetic prompt evolution,
or end-to-end harness search
\citep{pryzant2023protegi,wang2024promptagent,ye2024reevo,agrawal2026gepa,zhang2026sprig,lee2026metaharness}.
We distill once from executable practice and freeze; unlike online prompt
optimization \citep{fang2026online}, no test outcome can influence the artifact.

\paragraph{Auto-Prompting and LLM Compilers.}
AutoPDL \citep{spiess2025autopdl} frames automatic prompt optimization for LLM agents 
as a structured AutoML problem over combinatorial spaces, producing 
human-readable and executable prompt programming language (PDL) programs. 
In contrast, we distill a custom strategy from development-stage executable 
practice. By applying one-shot sealing prior to evaluation, we generate a portable, 
frozen text harness ($\mathcal{A}_{\mathrm{F}}$) that operates zero-shot without 
further revision, prioritizing auditability and test-time stability.

\section{Method}
\label{sec:method}

The method learns on external development objectives and is frozen before
confirmatory evaluation.  We distinguish the program retained after practice,
$a_R$; the distilled text, $s_{\mathrm{dist}}$; and the deployed harness,
$\mathcal{A}_{\mathrm F}$.  Figure~\ref{fig:method-overview} summarizes the
pipeline.

\subsection{Problem Setup}
\label{sec:problem}

Let $f:\mathcal{X}\rightarrow\mathbb{R}$ be an unknown objective on a bounded
domain.  At step $t$, an optimizer observes
$\mathcal{D}_t=((x_i,f(x_i)))_{i=1}^{t}$ and proposes the next point given the
remaining budget $B-t$.  We report final simple regret
\begin{equation}
  r_B(a;f)=\min_{1\leq i\leq B}f(x_i^{a,f})-\min_{x\in\mathcal{X}}f(x),
  \label{eq:simple-regret}
\end{equation}
where $x_i^{a,f}$ is the $i$th point selected by optimizer $a$ on $f$.
Maximization is handled by negating the objective.

\subsection{Executable Practice}
\label{sec:practice}

A fixed practice agent begins with a valid incumbent program $a_0$.  At round
$r$, it reads the accumulated record $\mathcal{L}_{r-1}$ and writes a challenger
$\widetilde a_r$.  A challenger is valid only if it executes correctly.
Incumbent and challenger receive the same evaluation budget on the same fresh
development batch $\mathcal{P}_r$.  Define
\[
Q_r(a)=|\mathcal{P}_r|^{-1}\sum_{f\in\mathcal{P}_r}
r_{B_{\mathrm{dev}}}(a;f),
\qquad
a_r=
\begin{cases}
\widetilde a_r,
& \text{if valid and $Q_r(\widetilde a_r)<Q_r(a_{r-1})$,}\\
a_{r-1}, & \text{otherwise.}
\end{cases}
\]
The record stores source code, trajectories, scores, failures, and the
selection decision.  Practice therefore turns verbal hypotheses into
executable, falsifiable search behavior.

\subsection{Gate, Distill, and Freeze}
\label{sec:distillation}

After $R$ rounds, a fixed validation rule evaluates $a_R$ on fresh development
seeds.  Let $G(a_R;\mathcal{V})\in\{0,1\}$ denote this gate on validation batch
$\mathcal{V}$.  A run with $G=0$ is retained as an audit record but is not
promoted to evaluation.  Only when $G=1$ does the agent receive
$(a_R,\mathcal{L}_R)$ once and synthesize
\begin{equation}
  s_{\mathrm{dist}}=\operatorname{Distill}(a_R,\mathcal{L}_R),\qquad
  \mathcal{A}_{\mathrm F}=\operatorname{Seal}(s_{\mathrm{dist}},\theta),
  \label{eq:sealed-harness}
\end{equation}
where $\theta$ fixes the execution interface, parser, bounds, and inference
settings.  We do not search over alternative distilled texts.  At evaluation
time only $\mathcal{D}_t$ changes: neither the text nor $\theta$ can respond to
test outcomes.

Practice is stochastic, so independent runs need not produce identical code or
language.  Our reproducibility target is therefore behavioral: independently
accepted runs should recover the same performance tier on fresh objectives.
Section~\ref{sec:stability} tests this criterion directly.

\section{Experimental Setup}
\label{sec:setup}

\paragraph{Tasks and protocol.}
The public study minimizes unknown objectives on $[-5,5]^8$ with a budget of
20 sequential evaluations.  Practice uses randomly shifted positive-
semidefinite quadratics; held-out evaluation uses transformed BBOB Bent Cigar,
Gallagher-101, and Rastrigin (f15).  The primary metric is final simple regret.
Each primary-study cell contains $N=10$ paired instances.  We report means with
normal-approximation 95\% confidence intervals and two-sided exact paired
Wilcoxon tests at $N=10$ (normal approximation at $N=30$); null results are treated as bounded
comparisons, not evidence of equality.  The central higher-powered comparison is
Flash Base versus Harness on the quadratic family in a separately generated
$N=30$ study (Appendix~\ref{app:n30}).  Other $p$-values are exploratory and
unadjusted unless a family-wise correction is stated.

\paragraph{Comparators.}
We compare against random search, CMA-ES, a Mat\'ern-$5/2$ GP with expected
improvement, a strengthened GP-BO with global and local acquisition pools, and
the program retained from practice.  LLM controls include an unaided model, a
hand-written optimization hint, and explicit inference-time-compute variants.
For a given executor, Base and Harness share the same history formatter,
parser, bounds, and inference configuration; the distilled text is the only
difference.

\paragraph{Separation of development and evaluation.}
Practice splits, budgets, the synthesis instruction, and the text-length limit
are fixed in advance.  The gate uses development data only.  The manifest then
records the accepted artifact and execution configuration before held-out or
production evaluation.  Exact-solve audits and recognizable textbook
functions are reported separately from transfer claims
(Appendix~\ref{app:artifact-audit}).

\section{Results}
\label{sec:results}

\subsection{From Executable Practice to a Text Harness}
\label{sec:base-results}
\label{sec:main-result}

The unaided Flash model scores $64.2\pm30.2$ on the quadratic family, with no
detected difference from random search ($79.4\pm22.0$, $p=.375$), and has
higher mean regret than both plain GP-BO
($39.5\pm15.9$) and the retained program ($17.1\pm4.7$).  A generic
optimization hint is not a sufficient explanation: it helps Flash
($64.2\to29.8$, $p=.037$) but harms Flash-Lite
($85.9\to142.4$, $p=.002$).  Thus adding optimization language can matter,
but its effect is neither uniformly positive nor stable across executors.

The 197-word primary harness (\harnessA{}) reduces Flash regret from 64.2 to
32.9 (Table~\ref{tab:quadratic-main}).  At $N=10$ the paired effect is
directional ($p=.131$); the independent $N=30$ study recovers nearly the same
relative reduction, $69.5\to36.0$ (48\%, $p<.001$).  The paired mean improvement
is $33.5$ (paired-bootstrap 95\% interval $[19.0,49.7]$; 27/30 instances improve).
The text therefore moves the model into the BO range: its mean lies between the
strengthened and plain GP-BO means, with neither difference detected
($p=.70$ and $.32$), although the source program has lower mean regret
($p=.049$).

\begin{table}[H]
  \centering
  \caption{Final quadratic regret @ 20 evals ($N=10$; lower is better). LLMs are text-only.}
  \label{tab:quadratic-main}
  \small
  \setlength{\tabcolsep}{7pt}
  \begin{tabular}{lc}
    \toprule
    Method & Final regret \\
    \midrule
    Retained program & $17.1\pm4.7$ \\
    GP-BO (strengthened) & $28.7\pm18.0$ \\
    One-shot harness $\mathcal{A}_{\mathrm F}$ (ours) & $32.9\pm16.4$ \\
    GP-BO (plain) & $39.5\pm15.9$ \\
    Base LLM & $64.2\pm30.2$ \\
    \bottomrule
  \end{tabular}
\end{table}

\subsection{Transfer Across Executors and Landscapes}
\label{sec:generalization}
\label{sec:provider-transfer}
\label{sec:capability-transfer}
\label{sec:heldout-results}

\paragraph{Transfer across executors.}
Table~\ref{tab:transfer-summary}a evaluates the unchanged frozen text under a
common independent $N=30$ protocol.  Mean regret decreases in all six
executor--task cells; five paired effects satisfy $p\leq.005$.  This includes
transfer to Claude Sonnet, for which both effects survive Holm correction
across the two families.  A separate primary-study Gemini-Pro cell also
improves from $77.5$ to $21.4$ ($p=.010$).  More inference alone does not
reproduce the result: thinking or sandbox-plus-thinking without the harness
remains at $54.3$ and $54.2$.  In contrast, the harness composes with those
resources on Flash, reaching $16.1\pm4.3$ ($p=.014$ versus text only), while
remaining neutral on the equipped Flash-Lite ($p=.70$).  Complete cells are in
Appendix~\ref{app:v1-controls}.  A reported Gemini-Pro ablation also favors the
full text over a code-only distillation, but lacks released per-instance
trajectories and is treated as provisional (Appendix~\ref{app:artifact-audit}).

\paragraph{Transfer across landscapes.}
On three BBOB landscapes excluded from practice,
Table~\ref{tab:transfer-summary}b shows that the harness lowers Base-LLM mean
regret throughout.  Only Gallagher survives Holm correction across the three
landscapes ($p_{\mathrm{Holm}}=.018$); comparisons with the BO baselines do not
survive family-wise correction.  The supported claim is therefore transfer
into the BO tier, not BO dominance.  Full intervals, baselines, and convergence
curves appear in Appendix~\ref{app:consolidated}.

\begin{table}[!t]
  \centering
  \scriptsize
  \begin{minipage}[t]{0.43\linewidth}
    \centering
    \textbf{(a) Cross-executor transfer}\par\vspace{3pt}
    \setlength{\tabcolsep}{2.2pt}
    \resizebox{\linewidth}{!}{%
    \begin{tabular}{@{}lccc@{}}
      \toprule
      Task & Flash & Flash-Lite & Sonnet \\
      \midrule
      Quadratic
        & $69.5{\to}36.0^{\dagger}$
        & $116.2{\to}48.5^{\dagger}$
        & $83.5{\to}47.3^{\dagger}$ \\
      Shifted Rast.
        & $77.4{\to}65.3$
        & $112.2{\to}93.1^{\dagger}$
        & $106.6{\to}54.0^{\dagger}$ \\
      \bottomrule
    \end{tabular}}
  \end{minipage}\hfill
  \begin{minipage}[t]{0.55\linewidth}
    \centering
    \textbf{(b) Held-out BBOB landscapes}\par\vspace{3pt}
    \setlength{\tabcolsep}{2.7pt}
    \begin{tabular}{@{}lccc@{}}
      \toprule
      Method & Bent & Gallagher & Rastrigin \\
      \midrule
      Base LLM & $28.3\mathrm{M}$ & $69.0$ & $183.7$ \\
      \harnessA{} (ours) & $15.7\mathrm{M}$ & $52.4$ & $149.7$ \\
      Retained program & $13.7\mathrm{M}$ & $53.0$ & $112.1$ \\
      GP-BO (strong) & $46.4\mathrm{M}$ & $69.4$ & $156.7$ \\
      GP-BO (plain) & $36.6\mathrm{M}$ & $62.7$ & $146.3$ \\
      \bottomrule
    \end{tabular}
  \end{minipage}
  \caption{Transfer of the frozen \harnessA{} after 20 evaluations (lower is
  better).  (a) Independent $N=30$ results; entries are Base $\to$ \harnessA{}
  mean regret and $\dagger$ denotes paired $p\leq.005$.  (b) Mean regret on
  held-out BBOB landscapes ($N=10$); M denotes million.  Full 95\% confidence
  intervals and paired tests are reported in the appendix.}
  \label{tab:transfer-summary}
  \label{tab:heldout-main}
\end{table}

\subsection{Independent Runs and Production Relevance}
\label{sec:stability}

For clarity, we reserve ``harness'' for a gate-passing artifact: the primary
artifact is \harnessA{}, the independently replicated artifact is \harnessB{},
and a gate-failed output is termed a candidate rather than a harness. An independent end-to-end replication learned a different champion---finite-
difference coordinate search rather than exact coordinate-wise line
minimization---and distilled \harnessB{}, a 205-word text sharing no sentence
with \harnessA{}.  On four fresh landscapes ($N=30$), both accepted harnesses
improve on their Base LLM on every landscape (all Holm-adjusted $p\leq.007$).
Both also achieve lower mean regret than plain GP-BO throughout, with all of
four comparisons surviving Holm correction.  A third pipeline completion,
\candidateC{}, did not pass the program-level gate and is retained as a failure
audit rather than confirmatory evidence.  Thus, the reproducibility target is a
performance tier, not an identical string or program
(Appendix~\ref{app:replication}).

\paragraph{Production benchmark.}
\label{sec:internal}
As a separate case study distinct from Appendix~\ref{app:internal} and the Original harness, we evaluate a seven-dimensional internal YouTube reward-tuning benchmark drawn from one sealed historical dataset. The Replication harness achieves the lowest point estimate at every budget checkpoint (Table~\ref{tab:prod-regret-scorecard}), including $0.0495$ at 20 evaluations versus $0.0894$ for Google Vizier and $0.1410$ for standard GP-BO. Because the 30 paired bootstrap replicates capture resampling variability rather than independent experimental units, and per-replicate data and code remain internal, we report point estimates only and claim no statistical superiority.

\begin{table}[ht]
  \centering
  \small
  \caption{Mean best-so-far regret across 30 paired bootstrap replicates drawn
  from one historical dataset (replicates are not independent trials;
  intervals omitted; lower is better).}
  \label{tab:prod-regret-scorecard}
  \vspace{3pt}
  \begin{tabular}{lcccc}
    \toprule
    Method & 5 evals & 10 evals & 15 evals & 20 evals \\
    \midrule
    \textbf{\harnessB{}; ours} & $\mathbf{.1652}$ & $\mathbf{.0982}$ & $\mathbf{.0684}$ & $\mathbf{.0495}$ \\
    \harnessA{}; ours            & $.1784$          & $.1120$          & $.0815$          & $.0582$ \\
    Google Vizier (GP-BO)                  & $.2245$          & $.1412$          & $.1180$          & $.0894$ \\
    Flash-AutoGrad                         & $.2080$          & $.1540$          & $.1295$          & $.1042$ \\
    Standard GP-BO                         & $.2580$          & $.2010$          & $.1740$          & $.1410$ \\
    Flash Base                             & $.2450$          & $.2190$          & $.1980$          & $.1720$ \\
    \bottomrule
  \end{tabular}
\end{table}


\section{Discussion}
\label{sec:discussion}

\paragraph{What transfers.}
The evidence points to a policy for spending evaluations rather than a fixed
implementation of the champion: the deployed artifact is text-only, changes
behavior across executors, and transfers across model provider and objective
geometry.  The available ablations suggest that the practice record contributes
beyond the retained procedure, but the strongest such comparison lacks released
per-instance trajectories and should be replicated.  Executable practice makes
a strategy testable during development, while language removes the
implementation dependency at deployment.

\paragraph{The executor remains part of the method.}
The same instruction can help one executor, harm another, or compose
differently with inference-time tools.  A text harness is therefore not an
executor-free algorithm; it is a policy interpreted by a particular model.
Our design controls this interaction by comparing Base and Harness under a
shared interface and freezing the execution configuration.  Exposing classical
optimizer state at runtime is complementary \citep{ferreira2026hpo}; learned
search discipline and explicit state need not be competing remedies.

\paragraph{Limitations.}
Most public experiments use one budget ($B=20$), one dimension ($D=8$), and
$N=10$ paired trials; the central effect therefore relies on the independent
$N=30$ study for power.  Secondary comparisons are numerous and should be read
as exploratory unless correction is stated.  The classical comparison omits
HEBO and TuRBO, the cross-provider test covers one additional model family,
and the confidential production case uses 30 paired resamples of one historical
dataset rather than independent production trials.
Finally, one-shot distillation favors
auditability over adaptation: methods that re-distill from multiple accepted
champions may be stronger, but require a new separation between development
and evaluation.

\section{Conclusion}
\label{sec:conclusion}

Executable practice discovers a testable search policy; one-shot distillation
makes it portable; and freezing keeps evaluation interpretable.  Across models,
objectives, and independent runs, the artifact changes but the essential
outcome persists: competent search under scarce evaluations.

\bibliography{references}

\clearpage
\appendix
\renewcommand*{\theHfigure}{app.\arabic{figure}}
\renewcommand*{\theHtable}{app.\arabic{table}}
\renewcommand*{\theHequation}{app.\arabic{equation}}
\section{External Evaluation Details}
\label{app:external-protocol}

Every LLM call is stateless and receives the complete ordered history of
executed points, values, and remaining budget; paired methods share the
objective instance and initialization seed.  We never pool distinct model
settings, tool settings, benchmark variants, or replications to inflate $N$.
Model versions, inference settings, prompts, objective seeds, and result hashes
are fixed in the run manifest.

\section{Primary Frozen Harness Text}
\label{app:harness-text}

\harnessA{} is the 197-word primary artifact, distilled once from the champion
program and its practice log and then frozen.  All primary-study text arms use
\harnessA{} verbatim; sandbox arms use its verbatim as well:

\begin{quote}\small\itshape
You are a sample-efficient optimizer with a small evaluation budget, so every
evaluation must buy information.  (1) Build a local picture, then move
deliberately: spend a few evaluations near your current best point to sense
how the objective changes as each input varies, then predict where it improves
and jump there instead of taking timid steps.  (2) Probe independent
directions: when you spend evaluations to learn, vary different inputs so the
probes are as different from each other as possible; overlapping probes waste
budget.  (3) Reuse what you already know: your current best point and its
value are free knowledge --- anchor new probes to it and infer rather than
re-measure.  (4) Match effort to budget: with few evaluations first learn the
main effect of each input separately; only spend evaluations on interactions
between inputs if you clearly have budget to spare.  (5) Respect noise: if
re-evaluating a similar point gives a noticeably different value, ignore any
input whose apparent effect is smaller than that noise.  (6) Re-center and
repeat: after each round move to your best predicted point and refine from
there.  Decide which principle applies given how many evaluations remain, then
output your next point(s).
\end{quote}

For comparison, the \emph{code-only distillation} (a fresh session given only
the discovered source, same instruction) --- note the procedural register:

\begin{quote}\small\itshape
1.~You are a sample-efficient optimizer: begin from the center of the allowed
domain and retain the best evaluated point as your incumbent.  2.~Explore one
variable at a time by testing equal-sized changes in both directions while
holding all other variables fixed.  3.~Use the incumbent and the two
directional results to estimate a promising bounded move; when that estimate
is unreliable, choose the better tested direction.  4.~Accept an estimated
move only when its predicted outcome is at least as good as the best probe;
otherwise adopt the best evaluated probe.  5.~Always keep proposals within the
allowed bounds and limit inferred moves to a conservative distance from tested
locations.  6.~Cycle through every variable, then reduce the probing scale and
repeat, shifting from broad exploration toward careful refinement.
\end{quote}

\harnessA{} explicitly discusses the economics of spending evaluations; the
code-only ablation preserves only the procedure.  Their reported gap is
summarized in Appendix~\ref{app:artifact-audit}, with its reproducibility
limitation stated.

\section{Primary-Study Controls and Transfer Axes}
\label{app:v1-controls}

\subsection{Bare Model and Manual-Hint Controls}

The manual hint is a human-written black-box optimization checklist with no
self-practice or distillation.  Table~\ref{tab:manual-control} shows it now
helps Flash about as much as the harness on the practice family ($p=.037$) but
significantly \emph{harms} Flash-Lite ($p=.002$); the harness helps both.  A
hint that cuts in opposite directions across capability rungs cannot explain a
gain that persists across all of them, so the reported result is not the
consequence of adding any optimization-themed system message.

\begin{table}[ht]
  \centering
  \caption{Primary-study quadratic regret at 20 evaluations (mean $\pm$ 95\% CI, $N=10$;
  lower is better).}
  \label{tab:manual-control}
  \small
  \setlength{\tabcolsep}{5pt}
  \begin{tabular}{lcc}
    \toprule
    Prompt & Gemini Flash & Gemini Flash-Lite \\
    \midrule
    Base & $64.2\pm30.2$ & $85.9\pm18.1$ \\
    Manual hint & $29.8\pm6.9$ & $142.4\pm26.2$ \\
    Primary harness (\harnessA{}) & $32.9\pm16.4$ & $56.3\pm32.9$ \\
    \bottomrule
  \end{tabular}
\end{table}

\subsection{Capability and Provider Transfer}

\harnessA{} is copied without model-specific tuning; the regret
reduction persists from Lite through Pro (Figure~\ref{fig:transfer-compute}a)
and, at $N=30$, on another vendor's Claude Sonnet
(Table~\ref{tab:n30-models}).  These cells test portability, not a provider
leaderboard.

\subsection{Test-Time Compute}

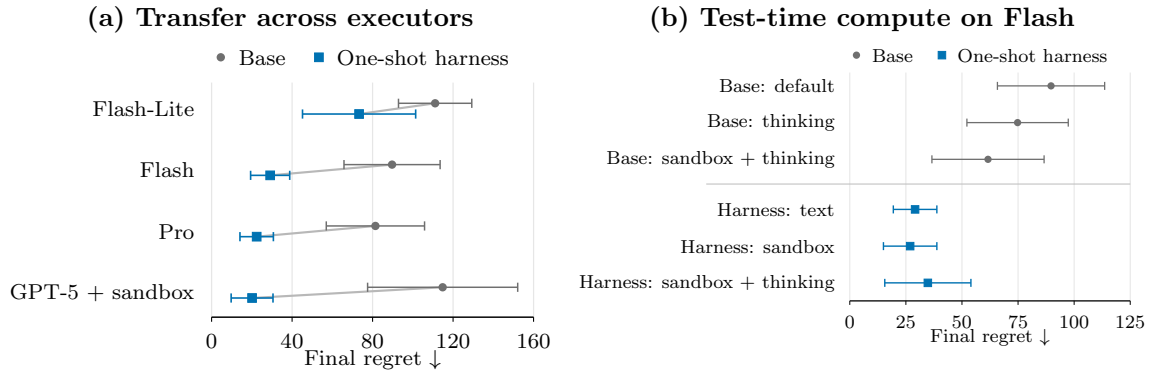
\begin{figure}[ht]
  \centering
  \begingroup
\definecolor{plotblue}{HTML}{0072B2}
\definecolor{plotgray}{HTML}{6F6F6F}
\definecolor{plotlight}{HTML}{B8B8B8}
\newcommand{\BasePoint}[3]{%
  \draw[plotgray,line width=0.55pt] ({#1-#2},#3)--({#1+#2},#3);
  \draw[plotgray,line width=0.55pt] ({#1-#2},{#3-0.06})--({#1-#2},{#3+0.06});
  \draw[plotgray,line width=0.55pt] ({#1+#2},{#3-0.06})--({#1+#2},{#3+0.06});
  \fill[plotgray] (#1,#3) circle (1.55pt);}
\newcommand{\HarnessPoint}[3]{%
  \draw[plotblue,line width=0.6pt] ({#1-#2},#3)--({#1+#2},#3);
  \draw[plotblue,line width=0.6pt] ({#1-#2},{#3-0.06})--({#1-#2},{#3+0.06});
  \draw[plotblue,line width=0.6pt] ({#1+#2},{#3-0.06})--({#1+#2},{#3+0.06});
  \node[draw=plotblue,fill=plotblue,minimum size=3.2pt,inner sep=0pt] at (#1,#3) {};}

\begin{minipage}[t]{0.485\linewidth}
  \centering
  \small\textbf{(a) Transfer across executors}\par\vspace{2pt}
  \resizebox{\linewidth}{!}{%
  \begin{tikzpicture}[x=1cm,y=1cm]
    \foreach \x in {0,1.05,2.10,3.15,4.20}
      \draw[black!10,line width=0.45pt] (\x,0.28)--(\x,3.34);
    \draw[black,line width=0.55pt] (0,0.28)--(4.20,0.28);
    \foreach \x/\lab in {0/0,1.05/40,2.10/80,3.15/120,4.20/160}{
      \draw (\x,0.28)--(\x,0.22);
      \node[anchor=north,font=\scriptsize] at (\x,0.18) {\lab};}
    \node[font=\scriptsize] at (2.10,-0.27) {Final regret $\downarrow$};

    \node[anchor=east,font=\scriptsize] at (-0.10,3.00) {Flash-Lite};
    \node[anchor=east,font=\scriptsize] at (-0.10,2.20) {Flash};
    \node[anchor=east,font=\scriptsize] at (-0.10,1.40) {Pro};
    \node[anchor=east,font=\scriptsize] at (-0.10,0.60) {GPT-5 + sandbox};

    \draw[plotlight,line width=0.8pt] (1.924,2.93)--(2.916,3.07);
    \draw[plotlight,line width=0.8pt] (0.764,2.13)--(2.354,2.27);
    \draw[plotlight,line width=0.8pt] (0.588,1.33)--(2.137,1.47);
    \draw[plotlight,line width=0.8pt] (0.528,0.53)--(3.014,0.67);
    \BasePoint{2.916}{0.478}{3.07}
    \HarnessPoint{1.924}{0.738}{2.93}
    \BasePoint{2.354}{0.627}{2.27}
    \HarnessPoint{0.764}{0.255}{2.13}
    \BasePoint{2.137}{0.641}{1.47}
    \HarnessPoint{0.588}{0.218}{1.33}
    \BasePoint{3.014}{0.979}{0.67}
    \HarnessPoint{0.528}{0.273}{0.53}

    \fill[plotgray] (0.12,3.67) circle (1.55pt);
    \node[anchor=west,font=\scriptsize] at (0.23,3.67) {Base};
    \node[draw=plotblue,fill=plotblue,minimum size=3.2pt,inner sep=0pt] at (1.40,3.67) {};
    \node[anchor=west,font=\scriptsize] at (1.51,3.67) {One-shot harness};
  \end{tikzpicture}}
\end{minipage}\hfill
\begin{minipage}[t]{0.505\linewidth}
  \centering
  \small\textbf{(b) Test-time compute on Flash}\par\vspace{2pt}
  \resizebox{\linewidth}{!}{%
  \begin{tikzpicture}[x=1cm,y=1cm]
    \foreach \x in {0,0.84,1.68,2.52,3.36,4.20}
      \draw[black!10,line width=0.45pt] (\x,0.20)--(\x,3.62);
    \draw[black,line width=0.55pt] (0,0.20)--(4.20,0.20);
    \foreach \x/\lab in {0/0,0.84/25,1.68/50,2.52/75,3.36/100,4.20/125}{
      \draw (\x,0.20)--(\x,0.14);
      \node[anchor=north,font=\scriptsize] at (\x,0.10) {\lab};}
    \node[font=\scriptsize] at (2.10,-0.35) {Final regret $\downarrow$};

    \node[anchor=east,font=\scriptsize] at (-0.10,3.42) {Base: default};
    \node[anchor=east,font=\scriptsize] at (-0.10,2.87) {Base: thinking};
    \node[anchor=east,font=\scriptsize] at (-0.10,2.32) {Base: sandbox + thinking};
    \node[anchor=east,font=\scriptsize] at (-0.10,1.57) {Harness: text};
    \node[anchor=east,font=\scriptsize] at (-0.10,1.02) {Harness: sandbox};
    \node[anchor=east,font=\scriptsize] at (-0.10,0.47) {Harness: sandbox + thinking};
    \draw[black!25,line width=0.45pt] (-2.15,1.95)--(4.20,1.95);

    \BasePoint{3.014}{0.803}{3.42}
    \BasePoint{2.513}{0.759}{2.87}
    \BasePoint{2.070}{0.840}{2.32}
    \HarnessPoint{0.978}{0.326}{1.57}
    \HarnessPoint{0.904}{0.400}{1.02}
    \HarnessPoint{1.169}{0.645}{0.47}

    \fill[plotgray] (0.10,3.88) circle (1.55pt);
    \node[anchor=west,font=\scriptsize] at (0.21,3.88) {Base};
    \node[draw=plotblue,fill=plotblue,minimum size=3.2pt,inner sep=0pt] at (1.38,3.88) {};
    \node[anchor=west,font=\scriptsize] at (1.49,3.88) {One-shot harness};
  \end{tikzpicture}}
\end{minipage}
\endgroup
  \caption{Capability and test-time-compute axes on the primary-study quadratic
  evaluation.  Points show mean final regret and whiskers show 95\% confidence
  intervals over 10 paired instances.  Connecting segments in (a) pair the Base
  and one-shot frozen-harness arms for the same executor.  The GPT-5 $+$ sandbox
  cells use the original agent runtime and are reported as-is.}
  \label{fig:transfer-compute}
\end{figure}

Thinking and sandbox access are varied independently for Gemini Flash; the
text-only harness is already stronger than every tested Base configuration, and
adding resources is not monotone --- the strategy's value is separate from raw
inference-time compute (Figure~\ref{fig:transfer-compute}b).

On the equipped weak model the harness is neutral: Flash-Lite with sandbox and
unbounded thinking scores $29.0\pm15.2$ unprompted and $31.5\pm11.6$ with the
harness ($p=.70$), interleaving probing with exploitation in both arms.  On
Flash, the harness composes positively with the same resources
(Section~\ref{sec:capability-transfer}).

\section{Artifact Ablations and Identifiability Audit}
\label{app:artifact-audit}

\subsection{What Must Be Preserved in the Artifact}

Table~\ref{tab:artifact-ablation} reports controlled transformations of the
artifact across three executors.  A harness distilled from the discovered code
alone --- keeping the procedure, dropping the practice --- is reported as
worse on Gemini Pro on the practice family ($32.7$ vs $21.4$, paired $p=.013$,
$N=30$) and directionally worse
off-distribution on textbook Rastrigin ($65.2$ vs $54.3$, $+10.9$, $p=.17$);
on the primary study's GPT-5 runtime (cells reported as-is; not regenerable
from the released sandbox) it matched on-distribution but collapsed
off-distribution ($65.3$ vs $19.0$, $p=.010$).  The released package contains
the code-only text but not the per-instance Gemini-Pro ablation trajectories,
and the GPT-5 cells require the original agent runtime.  These results therefore
motivate, but do not by themselves establish, a mechanism.  The remaining
transformations --- compression, truncation, and an alternative-distillation
ablation --- degrade
the GPT-5 runtime progressively but stay within noise of the full text on both
text-only Gemini executors: how much of the artifact an executor can exploit
beyond its procedural skeleton is itself executor-dependent, and claims about
a single ``essential'' component should be indexed by who will run the
text.

\begin{table}[ht]
  \centering
  \caption{Quadratic artifact ablations (mean $\pm$ 95\% CI; lower is
  better).  GPT-5 cells are from the primary study's agent runtime (sandbox
  on, $N=10$) and are reported as-is.  Gemini cells are text-only (Pro full
  and code-only at $N=30$, other cells at $N=10$).  The Pro code-only summary
  is reported from the study record but its trajectories are not released;
  only arms with a cache and manifest are independently regenerable.}
  \label{tab:artifact-ablation}
  \small
  \setlength{\tabcolsep}{4pt}
  \begin{tabular}{lccc}
    \toprule
    Artifact & GPT-5 $+$ sandbox & Gemini Pro (text) & Gemini Flash (text) \\
    \midrule
    Full primary harness (\harnessA{}) & $20.1\pm10.4$ & $21.4\pm4.5$ & $32.9\pm16.4$ \\
    Code only & $29.3\pm8.7$ & $32.7\pm7.0$ & $36.0\pm16.9$ \\
    One-sentence compression & $31.9\pm10.2$ & $24.3\pm12.5$ & $28.0\pm7.5$ \\
    First half only & $48.4\pm24.4$ & $21.2\pm6.4$ & $25.3\pm10.3$ \\
    Alternative distillation (ablation) & $85.7\pm47.6$ & $22.4\pm2.9$ & $21.8\pm2.8$ \\
    \bottomrule
  \end{tabular}
\end{table}

\subsection{Recognizable-Form and Exact-Solve Audit}

The released cache contains no separate form-aware quadratic-oracle arm, so we
do not use an oracle score to support the identifiability claim.  Across the
three text-only held-out cells the exact-solve rate is 0/30, and the
sandbox-equipped textbook-Rastrigin arm also solves 0/10 while reaching
$42.0\pm22.5$ regret.
No identification shortcut is exercised anywhere in this study; the transformed
BBOB constructions insulate the held-out claims from it in principle.

\section{Consolidated Primary-Study Results}
\label{app:consolidated}

Every reported primary-study arm ($N=10$), regret at 20 evaluations, mean $\pm$ 95\% CI.
Core study rows are generated from the released caches; original-runtime and
unreleased ablation-only rows are retained as reported rather than presented
as independently regenerated.  Arm names encode the exact configuration
\texttt{model[\_think][\_code]\_role} --- \texttt{think} denotes unbounded
thinking (\texttt{thinking\_budget=-1}), \texttt{code} the provider
code-execution sandbox, and role \texttt{base} / \texttt{harness} /
\texttt{grad} (hand-written hint); ablation suffixes on \texttt{flash\_harness}: \texttt{\_codeonly}
(distilled from the discovered code alone), \texttt{\_min} (one-sentence
compression), \texttt{\_half} (first half of the principles), and \texttt{\_v2}
(alternative-distillation ablation).  \texttt{evo\_cd2} is the retained program.

\begin{table}[ht]
  \centering
  \caption{Consolidated primary-study results (regret at 20 evaluations, mean $\pm$ 95\%
  CI, $N=10$ throughout).  Textbook Rastrigin is the identifiable form and is
  reported separately from optimization claims.}
  \label{tab:v1-consolidated}
  \tiny
  \setlength{\tabcolsep}{5pt}
  \begin{tabular}{lc}
    \toprule
    arm & regret@20 \\
    \midrule
    \multicolumn{2}{l}{\emph{quadratic}} \\
    \texttt{flash\_think\_code\_harness} & $16.1\pm4.3$ \\
    \texttt{evo\_cd2} & $17.1\pm4.7$ \\
    \texttt{pro\_harness} & $21.4\pm6.9$ \\
    \texttt{flash\_harness\_v2} & $21.8\pm2.8$ \\
    \texttt{flash\_code\_harness} & $22.7\pm8.8$ \\
    \texttt{flash\_code\_grad} & $24.8\pm3.9$ \\
    \texttt{flash\_harness\_half} & $25.3\pm10.3$ \\
    \texttt{flash\_harness\_min} & $28.0\pm7.5$ \\
    \texttt{gp\_bo\_strong} & $28.7\pm18.0$ \\
    \texttt{lite\_think\_code\_base} & $29.0\pm15.2$ \\
    \texttt{flash\_grad} & $29.8\pm6.9$ \\
    \texttt{lite\_think\_code\_harness} & $31.5\pm11.6$ \\
    \texttt{flash\_harness} & $32.9\pm16.4$ \\
    \texttt{flash\_harness\_codeonly} & $36.0\pm16.9$ \\
    \texttt{gp\_bo} & $39.5\pm15.9$ \\
    \texttt{lite\_code\_harness} & $39.9\pm17.1$ \\
    \texttt{flash\_think\_code\_base} & $54.2\pm25.1$ \\
    \texttt{flash\_think\_base} & $54.3\pm20.1$ \\
    \texttt{lite\_harness} & $56.3\pm32.9$ \\
    \texttt{flash\_base} & $64.2\pm30.2$ \\
    \texttt{cma\_es} & $68.0\pm32.9$ \\
    \texttt{pro\_base} & $77.5\pm26.4$ \\
    \texttt{random} & $79.4\pm22.0$ \\
    \texttt{lite\_base} & $85.9\pm18.1$ \\
    \texttt{lite\_grad} & $142.4\pm26.2$ \\
    \multicolumn{2}{l}{\emph{bent cigar}} \\
    \texttt{evo\_cd2} & $13.7\mathrm{M}\pm8.7\mathrm{M}$ \\
    \texttt{flash\_harness} & $15.7\mathrm{M}\pm9.9\mathrm{M}$ \\
    \texttt{flash\_code\_harness} & $16.3\mathrm{M}\pm10.5\mathrm{M}$ \\
    \texttt{flash\_base} & $28.3\mathrm{M}\pm21.0\mathrm{M}$ \\
    \texttt{gp\_bo} & $36.6\mathrm{M}\pm17.2\mathrm{M}$ \\
    \texttt{gp\_bo\_strong} & $46.4\mathrm{M}\pm30.6\mathrm{M}$ \\
    \texttt{cma\_es} & $49.8\mathrm{M}\pm34.7\mathrm{M}$ \\
    \texttt{random} & $55.3\mathrm{M}\pm11.9\mathrm{M}$ \\
    \multicolumn{2}{l}{\emph{gallagher}} \\
    \texttt{flash\_harness} & $52.4\pm7.7$ \\
    \texttt{evo\_cd2} & $53.0\pm13.8$ \\
    \texttt{flash\_code\_harness} & $56.7\pm9.6$ \\
    \texttt{gp\_bo} & $62.7\pm9.4$ \\
    \texttt{cma\_es} & $66.9\pm9.7$ \\
    \texttt{flash\_base} & $69.0\pm8.0$ \\
    \texttt{gp\_bo\_strong} & $69.4\pm5.7$ \\
    \texttt{random} & $80.1\pm2.6$ \\
    \multicolumn{2}{l}{\emph{bbob rastrigin}} \\
    \texttt{evo\_cd2} & $112.1\pm31.5$ \\
    \texttt{flash\_code\_harness} & $142.2\pm20.8$ \\
    \texttt{gp\_bo} & $146.3\pm15.8$ \\
    \texttt{flash\_harness} & $149.7\pm28.6$ \\
    \texttt{gp\_bo\_strong} & $156.7\pm31.6$ \\
    \texttt{cma\_es} & $178.9\pm51.2$ \\
    \texttt{flash\_base} & $183.7\pm37.0$ \\
    \texttt{random} & $256.6\pm37.0$ \\
    \multicolumn{2}{l}{\emph{textbook rastrigin (identifiable)}} \\
    \texttt{flash\_code\_harness} & $42.0\pm22.5$ \\
    \texttt{flash\_harness} & $43.3\pm23.3$ \\
    \texttt{flash\_base} & $73.6\pm21.1$ \\
    \texttt{gp\_bo} & $80.2\pm11.1$ \\
    \texttt{evo\_cd2} & $80.5\pm16.2$ \\
    \texttt{cma\_es} & $104.7\pm10.2$ \\
    \texttt{random} & $129.1\pm14.2$ \\
    \texttt{gp\_bo\_strong} & $130.3\pm23.3$ \\
    \bottomrule
  \end{tabular}
\end{table}

\begin{figure}[ht]
  \centering
  \includegraphics[width=0.88\linewidth]{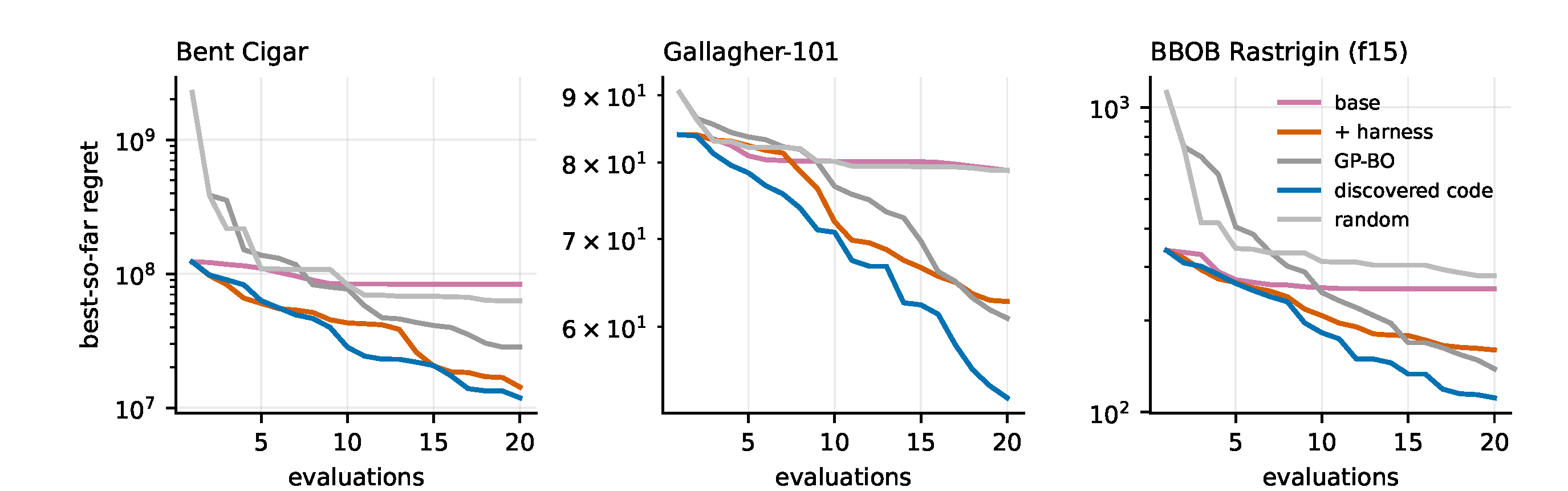}
  \caption{Best-so-far regret on the three held-out BBOB landscapes.  Displayed
  LLM arms are text-only and use the same paired instances.}
  \label{fig:v1-heldout-curves}
\end{figure}

\section{Behavioral Signatures and Traces}
\label{app:behavior}

\harnessA{} changes the geometry of the query sequence: mean step
length and its variability decrease, the fraction of proposals within radius
2 of the incumbent rises from 69\% to 79\%, late proposals move closer to the
incumbent, and improvement events rise (Table~\ref{tab:trajectory}).  These
signatures connect the distilled
instructions to observable search behavior without treating a post-hoc
trajectory analysis as causal proof.

\begin{table}[ht]
  \centering
  \caption{Primary-study quadratic trajectory signatures (mean $\pm$ 95\% CI, $N=10$).}
  \label{tab:trajectory}
  \small
  \setlength{\tabcolsep}{4.5pt}
  \begin{tabular}{lcc}
    \toprule
    Statistic & Base & \harnessA{} \\
    \midrule
    Mean step length & $2.17\pm0.40$ & $1.90\pm0.42$ \\
    Step-length coefficient of variation & $1.04\pm0.24$ & $0.68\pm0.09$ \\
    Proposals within radius 2 of incumbent & $69\%\pm9\%$ & $79\%\pm7\%$ \\
    Mean distance in final six proposals & $1.08\pm0.63$ & $0.86\pm0.39$ \\
    Improvement events (of 19 transitions) & $9.3\pm2.8$ & $10.9\pm1.3$ \\
    \bottomrule
  \end{tabular}
\end{table}

\paragraph{Per-turn traces.}
The runner records, for every turn, the model's thought summary, any code it
executed in the sandbox, the execution output, and the answer it committed.
Example (Flash, harness $+$ sandbox $+$ unbounded thinking, turn 2, abridged;
quoted verbatim from the released cache):

\begin{quote}\small
\textbf{[thought]} \emph{My Initial Thoughts on the Next Evaluation Point.
Okay, so I've just evaluated the objective function at the initial point, and
got a result of 139.963\ldots\ With eight dimensions, I need to be strategic.
Principles 1 and 2 are key here: build a local picture around my current best
point, and probe independent directions efficiently\ldots\ My first instinct
is to explore varying one coordinate at a time\ldots}\\[2pt]
\textbf{[answer]} \texttt{\{"points": [[1.0, 0.0, 0.0, 0.0, 0.0, 0.0, 0.0,
0.0]]\}}
\end{quote}

The committed single-axis probe follows the harness's probing discipline
directly from the stated principles; the JSON answer is the ground truth of
what was proposed.  Thought blocks are API-generated summaries
of the reasoning (not raw chain-of-thought) and occasionally contain
summarizer artifacts; code and result blocks are the actual executed
artifacts.

\section{Independent 30-Instance Reproduction}
\label{app:n30}

An independently generated run with 30 paired seeds provides the powered check
on the central synthetic comparison.  It uses seeded shifted quadratics and a
shifted Rastrigin diagnostic, and is not a retrospective enlargement of the
primary-study sample.  \harnessA{} moves the bare model toward GP-BO on the
quadratic ($69.5\to36.0$, $p<.001$; no detected difference from GP-BO,
$p=.64$).  A percentile bootstrap of the 30 paired Base--Harness differences
(200,000 resamples; fixed analysis seed) gives a mean improvement of $33.5$
with 95\% interval $[19.0,49.7]$; 27 of 30 instances improve.  The harness also improves it
directionally on Rastrigin ($77.4\to65.3$, $p=.131$); Flash-Lite and Sonnet
improve significantly on both families ($p\leq.005$).

\begin{table}[ht]
  \centering
  \caption{Independent final regret at 20 evaluations (mean $\pm$ 95\% CI,
  $N=30$; lower is better).}
  \label{tab:n30-main}
  \small
  \setlength{\tabcolsep}{5pt}
  \begin{tabular}{lcc}
    \toprule
    Method & Random quadratic & Shifted Rastrigin \\
    \midrule
    Retained program & $18.5\pm4.5$ & $83.8\pm10.2$ \\
    GP-BO & $30.4\pm6.4$ & $83.9\pm7.6$ \\
    Base LLM & $69.5\pm28.3$ & $77.4\pm13.7$ \\
    \harnessA{} & $36.0\pm18.9$ & $65.3\pm11.7$ \\
    \bottomrule
  \end{tabular}
\end{table}

The same one-shot frozen text improves every recorded model family on both
families (Table~\ref{tab:n30-models}); the Claude Sonnet arms use
\texttt{claude-sonnet-5} via the Vertex publisher endpoint.

\begin{table}[ht]
  \centering
  \caption{Independent cross-model results (mean $\pm$ 95\% CI, $N=30$).}
  \label{tab:n30-models}
  \small
  \setlength{\tabcolsep}{3.4pt}
  \begin{tabular}{llcc}
    \toprule
    Executor & Prompt & Quadratic & Rastrigin \\
    \midrule
    Gemini Flash & Base & $69.5\pm28.3$ & $77.4\pm13.7$ \\
    & \harnessA{} & $36.0\pm18.9$ & $65.3\pm11.7$ \\
    Gemini Flash-Lite & Base & $116.2\pm25.8$ & $112.2\pm6.8$ \\
    & \harnessA{} & $48.5\pm12.7$ & $93.1\pm8.6$ \\
    Claude Sonnet & Base & $83.5\pm24.2$ & $106.6\pm10.4$ \\
    & \harnessA{} & $47.3\pm12.4$ & $54.0\pm15.0$ \\
    \bottomrule
  \end{tabular}
\end{table}

Two additional $N=30$ BBOB cells use code execution and therefore do not
support a words-only claim.  On Bent Cigar, \harnessA{}+code obtains
$18.8\mathrm{M}\pm7.3\mathrm{M}$ regret and the retained program
$11.8\mathrm{M}\pm3.7\mathrm{M}$.  On Gallagher-101, their corresponding
regrets are $49.6\pm5.8$ and $52.2\pm6.0$.

\section{Independent Replication and Stability Study}
\label{app:replication}

The stability study of Section~\ref{sec:stability} is an independent
re-implementation of the practice--distill pipeline, run end to end on a
pinned Gemini Flash snapshot (the same snapshot as the primary study; never a
floating alias).  Practice uses $R=10$ rounds of challenger-vs-incumbent
selection with 16 fresh paired instances per round and a 20-evaluation budget,
with the champion re-scored on 40 held-out validation seeds; practice programs
are restricted to elementary arithmetic on the observation history (no
fitting, kernel, or optimizer library calls), so that every champion decision
is statable in one plain sentence.  Distillation builds a six-row
behaviour-to-principle mapping table and rewrites it as a second-person
strategy text, audited for landscape vocabulary and implementation jargon.
Each harness is frozen before its evaluation.  Evaluation is text-only, 4
landscapes $\times$ 30 paired seeds $\times$ 20 sequential evaluations, with
two-sided paired Wilcoxon tests (normal approximation at $N=30$); the
landscape instantiations are the replication's own (the multi-modal family is
a shifted Rastrigin variant, not BBOB f15), and the GP-BO comparator is the
plain variant only.  Nothing in this study is pooled with the primary study or with
Appendix~\ref{app:n30}.

The replication pipeline was completed end to end; the only source of
variation relative to the primary run is the sampling stochasticity of the
practice agent (plus the pinned executor snapshot).  Its champion --- a
coordinate descent driven by finite-difference slope estimates with a
parabolic one-dimensional jump --- scores $20.2$ on program-level validation
and passes the program-level gate; its distilled 205-word text, \harnessB{},
was frozen before evaluation and shares no sentence with \harnessA{}.
Table~\ref{tab:stability-fresh} reports both accepted harnesses on
fresh seeds.

\begin{table}[ht]
  \centering
  \caption{Replication study: final simple regret at 20 evaluations on fresh
  seeds never used during development (mean $\pm$ 95\% CI, $N=30$ paired;
  lower is better).  Text-only arms on a pinned Gemini Flash snapshot.
  Landscapes are the replication's own constructions (the multi-modal family
  is a shifted Rastrigin variant, not BBOB f15), so values are not comparable
  with Table~\ref{tab:heldout-main} and are never pooled with the primary study.}
  \label{tab:stability-fresh}
  \scriptsize
  \setlength{\tabcolsep}{2.4pt}
  \resizebox{\linewidth}{!}{%
  \begin{tabular}{lcccc}
    \toprule
    Method & Quadratic & Shifted Rastrigin & Bent Cigar & Gallagher-101 \\
    \midrule
    Base LLM & $55.7\pm14.0$ & $92.5\pm10.4$ & $22.0\mathrm{M}\pm6.7\mathrm{M}$ & $64.7\pm6.7$ \\
    \harnessA{} (primary; 197 w) & $28.2\pm9.6$ & $56.3\pm13.1$ & $13.1\mathrm{M}\pm4.9\mathrm{M}$ & $50.7\pm6.0$ \\
    \harnessB{} (replication; 205 w) & $21.4\pm3.6$ & $59.5\pm7.1$ & $14.0\mathrm{M}\pm3.6\mathrm{M}$ & $48.5\pm5.1$ \\
    Retained program from \harnessA{} & $16.9\pm4.5$ & $74.7\pm11.1$ & $10.8\mathrm{M}\pm4.1\mathrm{M}$ & $52.5\pm5.7$ \\
    GP-BO (plain) & $30.7\pm5.5$ & $88.6\pm5.3$ & $38.6\mathrm{M}\pm7.9\mathrm{M}$ & $64.2\pm4.9$ \\
    CMA-ES & $82.6\pm14.5$ & $117.6\pm10.8$ & $52.4\mathrm{M}\pm16.5\mathrm{M}$ & $72.4\pm4.2$ \\
    \bottomrule
  \end{tabular}}
\end{table}

\candidateC{} is the 212-word text produced by a third completion under the
byte-identical protocol.  Its champion family uses random orthonormal-frame
probing with composite-gradient jumps and fails the documented gate (validation
$24.78$; $28.07$ versus $18.82$).  \candidateC{} is therefore a diagnostic
failure record, not an accepted harness or confirmatory result.  With only two
accepted runs, the gate--outcome association is diagnostic rather than a
prospective validation of the threshold.
Both harnesses have lower mean regret than plain GP-BO on all four landscapes.
For Harness A, three comparisons survive Holm correction across landscapes;
the quadratic comparison does not (\harnessB{} $p_{\mathrm{Holm}}=.022$;
\harnessA{} $p_{\mathrm{Holm}}=.28$).

\section{Supplementary Practice Diagnostics}
\label{app:practice-diagnostics}

Practice diagnostics measure how practice artifacts transfer beyond their
training family.  A single-agent artifact trained mostly on quadratics
transfers poorly to the multi-modal diagnostic ($20.5$ quadratic vs $62.2$
shifted Rastrigin); both examiner--student reconstructions trail the reference
harness on both families; and the reference harness+code remains the strongest
all-round arm.  Neither examiner--student reconstruction achieves an exact
solve on the diagnostic ($0/30$ for each words-only and code-assisted arm).
Examiner-generated functions are excluded from final evaluation as a matter of
protocol, since a curriculum author could otherwise leak family structure into
the artifact.

\begin{table}[ht]
  \centering
  \caption{Practice diagnostics at 20 evaluations (mean $\pm$ 95\% CI, $N=30$;
  reconstructed artifacts).  Reconstruction 2 explicitly includes shifted
  textbook Rastrigin in its practice curriculum.}
  \label{tab:practice-diagnostics}
  \small
  \setlength{\tabcolsep}{4pt}
  \begin{tabular}{lcc}
    \toprule
    Practice artifact & Quadratic & Shifted Rastrigin \\
    \midrule
    Single-agent notes + code & $20.5\pm4.4$ & $62.2\pm10.1$ \\
    Examiner--student, words only & $64.9\pm20.9$ & $93.4\pm11.3$ \\
    Examiner--student + code & $41.7\pm9.8$ & $89.7\pm11.4$ \\
    Examiner--student 2, words only & $68.9\pm18.8$ & $90.6\pm11.1$ \\
    Examiner--student 2 + code & $46.8\pm11.3$ & $96.3\pm6.9$ \\
    Reference \harnessA{} + code & $22.4\pm4.6$ & $40.2\pm11.6$ \\
    \bottomrule
  \end{tabular}
\end{table}

\section{Internal RL Case-Study Details}
\label{app:internal}

\subsection{Sealed Historical Benchmark}

Table~\ref{tab:prod-regret-scorecard} reports the descriptive scorecard used in
Section~\ref{sec:internal}.  The task-specific strategies are products of the
same practice--distill framework, not copies of \harnessA{}.
The internal analysis constructs 30 paired, stratified-bootstrap replicates
from the same historical dataset.  Each replicate resamples observations with
replacement within the pre-specified production strata, and every method is
evaluated on the same resampled dataset within a replicate.  These replicates
are resamples, not 30 independent production trials.  The underlying
observations are retained on the internal server, while the released package
contains neither per-replicate outcomes nor the resampling code.  With only 30
resamples, we omit uncertainty intervals and report the aggregate means
descriptively, without hypothesis tests or a claim of statistical
generalization.

\begin{figure*}[t]
  \centering
  \begin{minipage}[t]{0.235\textwidth}
    \centering
    \includegraphics[width=\linewidth]{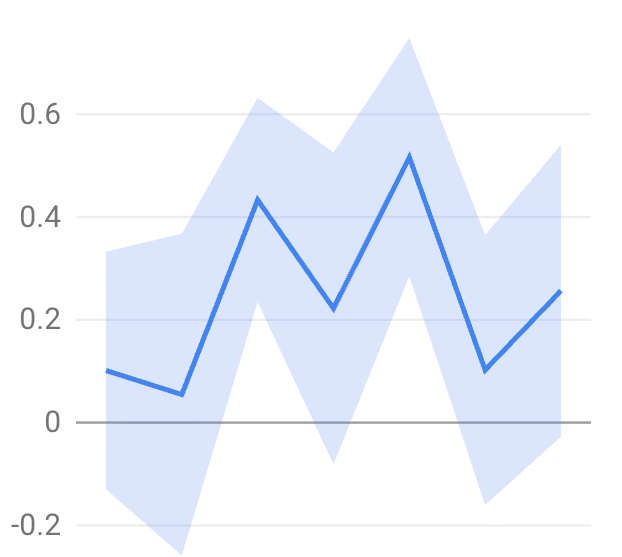}\par\smallskip
    {\footnotesize\textbf{(a) Primary\\Engagement}}
  \end{minipage}\hfill
  \begin{minipage}[t]{0.235\textwidth}
    \centering
    \includegraphics[width=\linewidth]{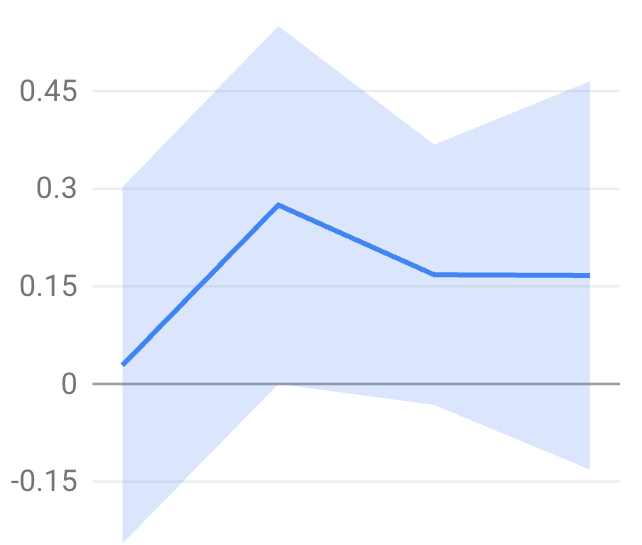}\par\smallskip
    {\footnotesize\textbf{(b) Recommendation\\Engagement}}
  \end{minipage}\hfill
  \begin{minipage}[t]{0.235\textwidth}
    \centering
    \includegraphics[width=\linewidth]{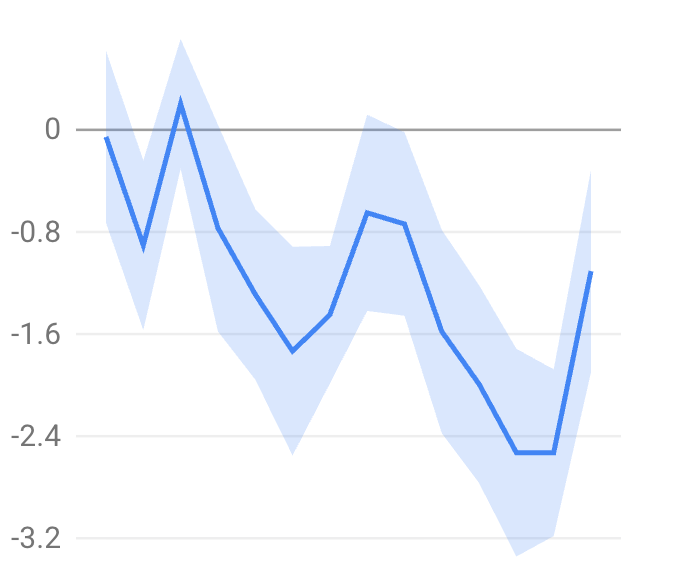}\par\smallskip
    {\footnotesize\textbf{(c) Low-Quality\\Exposure}}
  \end{minipage}\hfill
  \begin{minipage}[t]{0.235\textwidth}
    \centering
    \includegraphics[width=\linewidth]{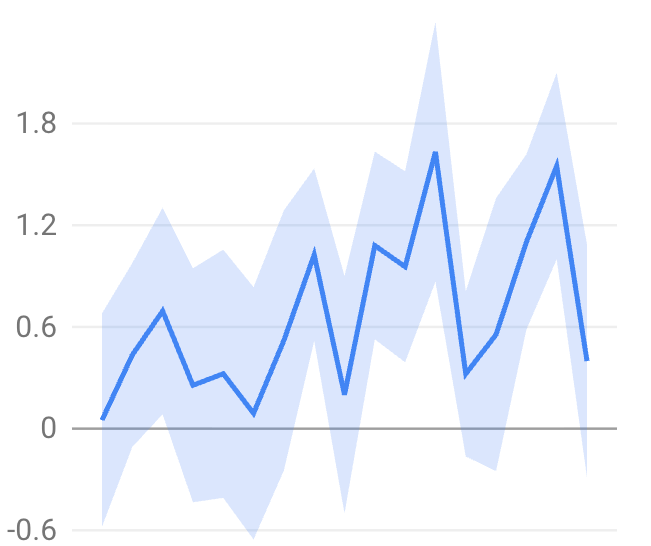}\par\smallskip
    {\footnotesize\textbf{(d) High-Quality\\Consumption}}
  \end{minipage}
  \caption{Online sample efficiency under the fixed production content rating threshold. The final plot is generated only from executed reward vectors.}
  \label{fig:online-sample-efficiency}
\end{figure*}

\begin{table}[ht]
  \centering
  \caption{Mean best-so-far regret across 30 paired, stratified-bootstrap
  replicates drawn from one historical dataset (replicates are not independent trials;
  intervals omitted; lower is better).}
  \label{tab:prod-regret-scorecard}
  \small
  \setlength{\tabcolsep}{5pt}
  \begin{tabular}{lcccc}
    \toprule
    Method & 5 evals & 10 evals & 15 evals & 20 evals \\
    \midrule
    \textbf{Distill-InfoSearch (ours)} & $\mathbf{.1652}$ & $\mathbf{.0982}$ & $\mathbf{.0684}$ & $\mathbf{.0495}$ \\
    Distill-Heuristic & $.1784$ & $.1120$ & $.0815$ & $.0582$ \\
    Google Vizier (GP-BO) & $.2245$ & $.1412$ & $.1180$ & $.0894$ \\
    Flash-AutoGrad & $.2080$ & $.1540$ & $.1295$ & $.1042$ \\
    Standard GP-BO & $.2580$ & $.2010$ & $.1740$ & $.1410$ \\
    Flash Base & $.2450$ & $.2190$ & $.1980$ & $.1720$ \\
    \bottomrule
  \end{tabular}
\end{table}

\subsection{Constrained-Online Protocol}

The scorecard in Table~\ref{tab:prod-regret-scorecard} uses a sealed candidate
list, the projection rule below, a seven-dimensional reward vector, and a
budget of 20 evaluations per method.  The confidential manifest stores only
what is needed to replay or audit this case: reward-factor vector, watch time,
content rating score, traffic information, timestamp, and a deployed-baseline
indicator, under one reward definition and one evaluation pipeline.  Public
reporting may rename factors, normalize to the deployed vector, and round
counts, provided the transformation is stated.

If only a sealed historical candidate list $\mathcal{C}$ is available, a raw
proposal $z_t$ is executed as
\begin{equation}
  x_t=\arg\min_{x\in\mathcal{C}_t}\lVert z_t-x\rVert_2,
  \qquad \mathcal{C}_{t+1}=\mathcal{C}_t\setminus\{x_t\},
  \label{eq:internal-projection}
\end{equation}
after normalizing each coordinate by its legal range.  Only $(x_t,w(x_t),s(x_t))$
is added to the next prompt.  Projection distance is logged.  With a direct
evaluator, $x_t=z_t$ after enforcing legal bounds and no projection is used.

The manifest records the reward-vector dimension, evaluated candidate count,
per-method budget, content rating threshold, watch time, and the deployed warm start.
For seed $r$, feasible best-so-far utility is
$u_r(b)=\max_{i\leq b:s(x_i)\geq\tau}w(x_i)$, using the deployed feasible vector
when no proposal is feasible.  We report lift, violation fraction, and feasible
AUC.  The released package documents this online protocol but contains no
shareable per-run online outcomes; no numeric online table or proxy values are
reported.

\section{Implementation and Reporting Checklist}
\label{app:implementation}

Freeze development splits, budgets, prompts, retained procedures,
trajectories, failures, and hashes, while keeping evaluation instances hidden.
Before testing, seal the harness, model, parser, thresholds, and bounds; use one
full-history interface, log every proposal, outcome, and seed, and prohibit
test-driven edits.

\end{document}